\documentclass[oribibl]{llncs}  
\usepackage[width=122mm,left=12mm,paperwidth=146mm,height=193mm,top=12mm,paperheight=217mm]{geometry}
\usepackage[T1]{fontenc}
\usepackage[utf8]{inputenc}
\usepackage{authblk}
\usepackage{etoolbox} 
\usepackage[dvipsnames]{xcolor}

\usepackage{times}
\usepackage{epsfig}
\usepackage{graphicx}
\usepackage{amsmath}
\usepackage{amssymb}
\usepackage{float}
\usepackage{color}
\usepackage{hyperref}

\usepackage[justification=centering]{subfig}
\usepackage[ruled,vlined]{algorithm2e}

\begin{document}
\title{Contextual Residual Aggregation for Ultra High-Resolution Image Inpainting}

\author{Zili Yi\qquad Qiang Tang\qquad Shekoofeh Azizi\qquad Daesik Jang\qquad Zhan Xu}


\institute{	Huawei Technologies Canada Co. Ltd.} 

\maketitle

\begin{figure}
\begin{center}
  \includegraphics[width=\linewidth]{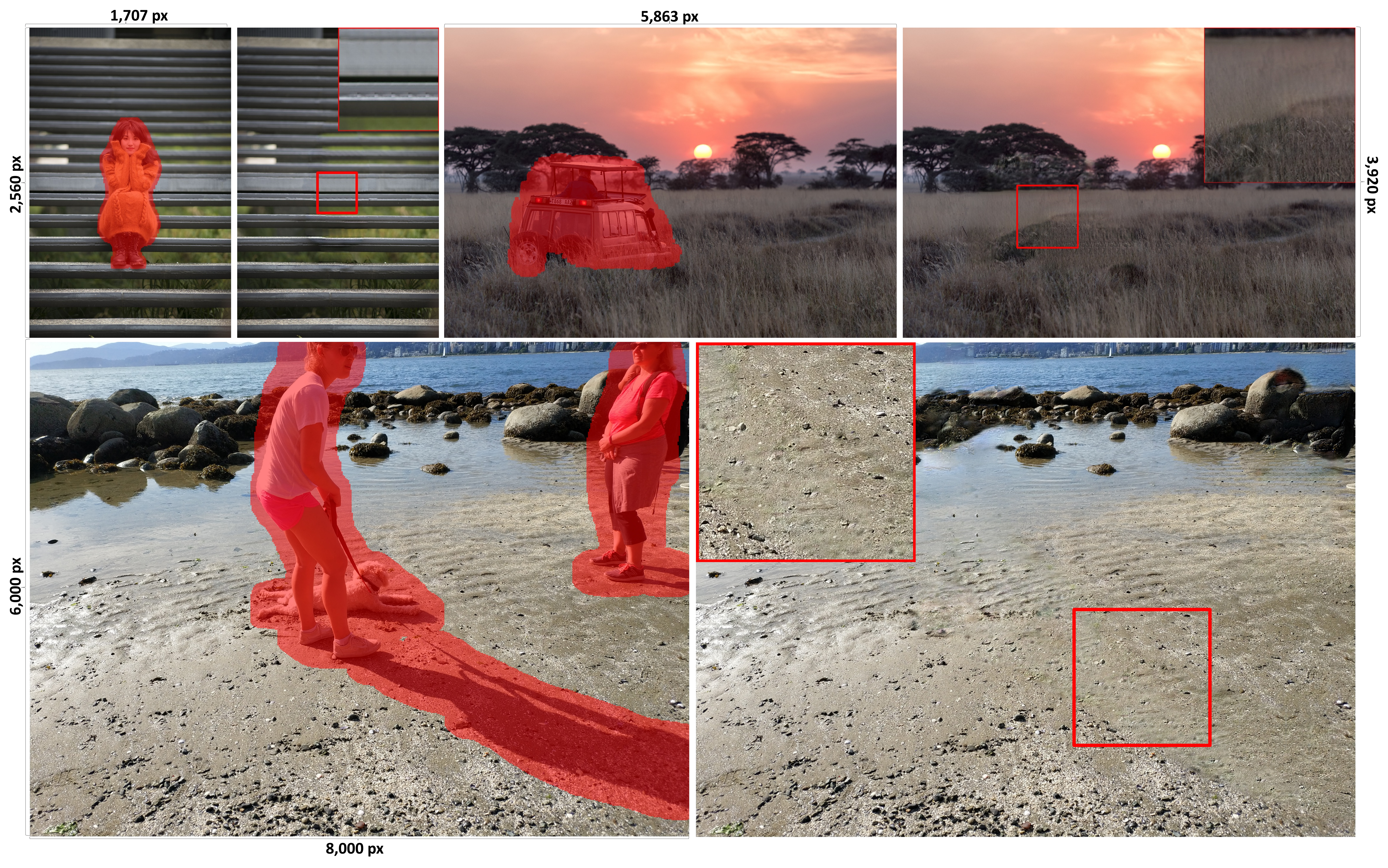} 
	\captionof{figure}{Inpainting results on ultra high-resolution images.\label{fig:hd3}}
\end{center}
\end{figure}

\begin{abstract}
Recently data-driven image inpainting methods have made inspiring progress, impacting fundamental image editing tasks such as object removal and damaged image repairing. These methods are more effective than classic approaches,  however, due to memory limitations they can only handle low-resolution inputs, typically smaller than 1K. Meanwhile, the resolution of photos captured with mobile devices increases up to 8K. Naive up-sampling of the low-resolution inpainted result can merely yield a large yet blurry result. Whereas, adding a high-frequency residual image onto the large blurry image can generate a sharp result, rich in details and textures. Motivated by this, we propose a Contextual Residual Aggregation (CRA) mechanism that can produce high-frequency residuals for missing contents by weighted aggregating residuals from contextual patches, thus only requiring a low-resolution prediction from the network. Since convolutional layers of the neural network only need to operate on low-resolution inputs and outputs, the cost of memory and computing power is thus well suppressed. Moreover, the need for high-resolution training datasets is alleviated. In our experiments, we train the proposed model on small images with resolutions 512$\times$512 and perform inference on high-resolution images, achieving compelling inpainting quality. Our model can inpaint images as large as 8K with considerable hole sizes, which is intractable with previous learning-based approaches. We further elaborate on the light-weight design of the network architecture, achieving real-time performance on 2K images on a GTX 1080 Ti GPU. Codes are available at: \href{https://github.com/Atlas200dk/sample-imageinpainting-HiFill}{Atlas200dk/sample-imageinpainting-HiFill}.
\end{abstract}

\section{Introduction}
Smartphone users are interested to manipulate their photographs in any form of altering object positions, removing unwanted visual elements, or repairing damaged images. These tasks require automated image inpainting, which aims at restoring lost or deteriorated parts of an image given a corresponding mask. Inpainting has been an active research area for the past few decades, however, due to its inherent ambiguity and the complexity of natural images, general image inpainting remains challenging. High-quality inpainting usually requires generating visually realistic and semantically coherent content to fill the hole regions. Existing methods for image hole filling can be categorized into three groups. The first category which we call~\emph{``fill through copying''} attempts to explicitly borrow contents or textures from surroundings to fill the missing regions. An example is diffusion-based~\cite{ballester2001filling,bertalmio2000image} methods which propagate local image appearance surrounding the target holes based on the isophote direction field. Another stream is relying on texture synthesis techniques, which fills the hole by both extending and borrowing textures from surrounding regions~\cite{criminisi2004region, drori2003fragment, he2012statistics, wilczkowiak2005hole, xu2010image}. Patch-based algorithms like~\cite{drori2003fragment,efros2001image,efros1999texture,wilczkowiak2005hole} progressively fill pixels in the hole by searching the image patches from background regions that are the most similar to the pixels along the hole boundaries.

The second group attempts to~\emph{``fill through modeling''} and hallucinates missing pixels in a data-driven manner with the use of large external databases. These approaches learn to model the distribution of the training images and assume that regions surrounded by similar contexts likely to possess similar contents~\cite{iizuka2017globally, liao2018edge, oord2016pixel, pathak2016context, xiong2019foreground, yang2017high}. For instance, PixelRNN~\cite{oord2016pixel} uses a two-dimensional Recurrent Neural Network (RNN) to model the pixel-level dependencies along two spatial dimensions. More general idea~\cite{iizuka2017globally,yang2017high} is to train an encoder-decoder convolutional network to model the 2-dimensional spatial contents. Rather than modeling the raw pixels,~\cite{liao2018edge, xiong2019foreground} train a convolutional network to model image-wide edge structure or foreground object contours, thus enabling auto-completion of the edge or contours. These techniques are effective when they find an example image with sufficient visual similarity to the query, but will easily fail if the database does not have similar examples. To overcome the limitation of copying-based or modeling-based methods, the third group of approaches attempts to combine the two ~\cite{liu2018image,song2018contextual,yan2018shift,yu2018free,yu2018generative,zeng2019learning}. These methods learn to model the image distribution in a data-driven manner, and in the meantime, they develop mechanisms to explicitly borrow patches/features from background regions.~\cite{yu2018generative} introduces a novel contextual attention layer that enables borrowing features from distant spatial locations.~\cite{zeng2019learning} further extends the contextual attention mechanism to multiple scales and all the way from feature-level to image-level.~\cite{song2018contextual} employs the patch-swap layer that propagates the high-frequency texture details from the boundaries to hole regions.

Most learning-based approaches belong to the second or third group. Compared to traditional methods, these techniques have strong ability to learn adaptive and high-level features of disparate semantics and thus are more adept in hallucinating visually plausible contents especially when inpainting structured images like faces~\cite{iizuka2017globally,oord2016pixel,song2018contextual,yu2018free,yu2018generative,zeng2019learning}, objects~\cite{liao2018edge,pathak2016context,xiong2019foreground,yang2017high}, and natural scenes~\cite{iizuka2017globally,song2018contextual,yu2018free,yu2018generative}. Since existing methods employ convolutional layers directly on the original input, the memory usage could become extremely high and intractable when the input size is up to 8K. Another issue is that the quality deteriorates rapidly when hole size increases with image size. Even if the training is feasible, access to large amounts of high-resolution training data would be tedious and expensive.

To resolve these issues, we propose a novel Contextual Residual Aggregation (CRA) mechanism to enable the completion of ultra high-resolution images with limited resources. In specific, we use a neural network to predict a low-resolution inpainted result and up-sample it to yield a large blurry image. Then we produce the high-frequency residuals for in-hole patches by aggregating weighted high-frequency residuals from contextual patches. Finally, we add the aggregated residuals to the large blurry image to obtain a sharp result. Since the network only operates on low-resolution images, the cost of memory and computing time is significantly reduced. Moreover, as the model can be trained with low-resolution images, the need for high-resolution training datasets is alleviated. Furthermore, we introduce other techniques including slim and deep layer configuration, attention score sharing, multi-scale attention transfer, and Light-Weight Gated Convolutions (LWGC) to improve the inpainting quality, computation, and speed. Our method can inpaint images as large as 8K with satisfying quality, which cannot be handled by prior learning-based approaches. Exemplar results are shown in Figure \ref{fig:hd3}. The contributions of the paper are summarized as follows:
\begin{itemize}
	\item[--]{We design a novel and efficient Contextual Residual Aggregation (CRA) mechanism that enables ultra high-resolution inpainting with satisfying quality. The mechanism enables large images (up to 8K) with considerable hole sizes (up to 25\%) to be inpainted with limited memory and computing resources, which is intractable for prior methods. Also, the model can be trained on small images and applied on large images, which significantly alleviates the requirements for high-resolution training datasets.}
	\item[--]{We develop a light-weight model for irregular hole-filling that can perform real-time inference on images of 2K resolutions on a NVIDIA GTX 1080 Ti GPU, using techniques including slim and deep layer configuration, attention score sharing, and Light Weight Gated Convolution (LWGC).}
	\item[--]{We use attention transfer at multiple abstraction levels which enables filling holes by weighted copying features from contexts at multiple scales, improving the inpainting quality over existing methods by a certain margin even when tested on low-resolution images.}
\end{itemize}

\begin{figure*}[t]
\begin{center}
  \includegraphics[width=\linewidth]{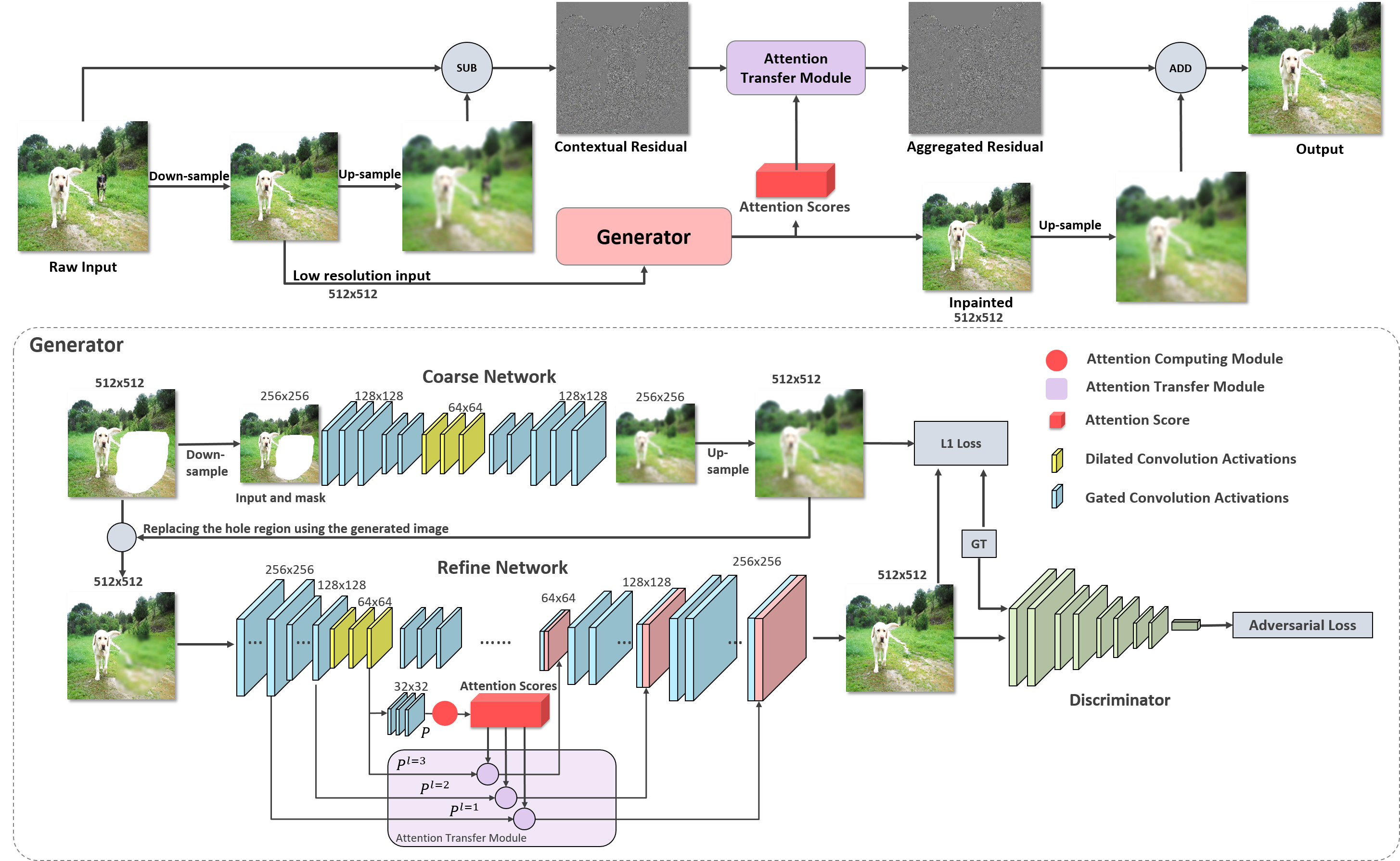}
\end{center}
   \caption{The overall pipeline of the method: (top) CRA mechanism, (bottom) the architecture of the generator.}
\label{fig:method}
\end{figure*}

\section{Related Works}
\subsection{Irregular Hole-filling \& Modified Convolutions}
Vanilla convolutions are intrinsically troublesome for irregular hole-filling because convolutional filters treat all pixels the same as valid ones, causing visual artifacts such as color inconsistency, blurriness, and boundary artifacts. Partial convolution~\cite{liu2018image} is proposed to handle irregular holes, where the convolution is masked and re-normalized to be conditioned on valid pixels. Gated convolution~\cite{yu2018free} generalizes the partial convolution idea by providing a learnable dynamic feature selection mechanism for each channel and at each spatial location, achieving better visual performance. Here, we further improve the gated convolution through a lightweight design to improve efficiency.

\subsection{Contextual Attention}
Contextual attention~\cite{yu2018generative} is proposed to allow long-range spatial dependencies during inpainting, which enables borrowing pixels from distant locations to fill missing regions. The contextual attention layer has two phases: ``match'' and ``attend''. In the ``match'' phase, the attention scores are computed by obtaining region affinity between patches inside and those outside the holes. In the ``attend'' phase, holes are filled by copying and aggregating patches from weighted contexts by the attention scores.~\cite{zeng2019learning} extends this idea by using a pyramid of contextual attention at multiple layers. In contrast to~\cite{zeng2019learning}, we only compute the attention scores once and reuse them at multiple abstraction levels, which leads to fewer parameters and less computation. 

\subsection{Image Residuals}
The difference between an image and the blurred version of itself represents the high-frequency image~\cite{burt1983laplacian,denton2015deep}. Early works use the difference obtained by Gaussian blurring for low-level image processing tasks like edge detection, image quality assessment, and feature extraction~\cite{deshmukh2010image,sharifi2002classified,toet1989image}. We employ this concept to decompose the input image into low-frequency and high-frequency components. The low-frequency component is obtained through averaging neighboring pixels, whereas the high-frequency component (i.e. image residuals) is obtained by subtracting the original image with its low-frequency component.

\section{Method}
\subsection{The Overall Pipeline}
Figure~\ref{fig:method} illustrates the overall pipeline of the proposed CRA mechanism where the generator is the only trainable component in the framework. Given a high-resolution input image, we first down-sample the image to $512\times512$ and then up-sample it to obtain a blurry large image of the same size as the raw input (Section~\ref{sect:cra}). The height and width of the image are not necessary to be equal but must be multiples of 512. The generator takes the low-resolution image and fills the holes. Meanwhile, the attention scores are calculated by the Attention Computing Module (ACM) of the generator (Section~\ref{sect:acm}). Also, the contextual residuals are computed by subtracting the large blurry image from the raw input, and the aggregated residuals in the mask region are then calculated from the contextual residuals and attention scores through an Attention Transfer Module (ATM) (Section~\ref{sect:atm}). Finally, adding the aggregated residuals to the up-sampled inpainted result generates a large sharp output in the mask region while the area outside mask is simply a copy of the original raw input.

\subsection{Contextual Residual Aggregation (CRA)}
Filling the missing region by using contextual information~\cite{song2018contextual,yan2018shift,yu2015multi}, and contextual attention mechanism~\cite{yu2018generative} has been proposed previously. Similarly, we deploy the CRA mechanism to borrow information from contextual regions. However, the CRA mechanism borrows from contexts not only features but also residuals. In particular, we adopt the idea of contextual attention~\cite{yu2018generative} in calculating attention scores by obtaining region affinity between patches inside/outside missing regions. Thus contextually relevant features and residuals outside can be transferred into the hole. Our mechanism involves two key modules: Attention Computing Module and Attention Transfer Module.

\subsubsection{Attention Computing Module (ACM)}
\label{sect:acm}
The attention scores are calculated based on region affinity from a high-level feature map (denoted as $P$ in Figure~\ref{fig:method}). $P$  is divided into patches and ACM calculates the cosine similarity between patches inside and outside missing regions:

\begin{equation}
  c_{i,j}  = \big < \frac{p_i}{\|p_i\|}, \frac{p_j}{\|p_j\|} \big >\\
\end{equation}

\noindent where $p_i$ is the $i^{th}$ patch extracted from $P$ outside mask, $p_j$ is the $j^{th}$ patch extracted from $P$ inside the mask. Then softmax is applied on the similarity scores to obtain the attention scores for each patch:

\begin{equation}
  s_{i,j} = \frac{e^{c_{i,j}}}{\Sigma_{i=1}^N {e^{c_{i,j}}}}\\
\end{equation}

\noindent where $N$ is the number of patches outside the missing hole. In our framework, each patch size is $3\times3$ and $P$ is $32\times32$, thus a total number of $1024$ patches can be extracted. In practice, the number of in-hole patches could vary for different hole sizes. We uniformly use a matrix of $1024\times1024$ to save affinity scores between any possible pair of patches, though only a fraction of them are useful.

\subsubsection{Attention Transfer Module (ATM)}\label{sect:atm}
After obtaining the attention scores from $P$, the corresponding holes in the lower-level feature maps ($P^l$) can be filled with contextual patches weighted by the attention scores:

\begin{equation}
 p^l_j = \Sigma_{i=1}^N { s_{i,j} p^l_i}\\
\end{equation}

\noindent where $l \in {1,2,3}$ is the layer number and $p^l_i$ is the $i^{th}$ patch extracted from $P^l$ outside masked regions, and $p^l_j$ is the $j^{th}$ patch to be filled inside masked regions. $N$ indicates the number of contextual patches (background). After calculating all in-hole patches, we can finally obtain a filled feature $P^l$. As the size of feature maps varies by layer, the size of patches should vary accordingly. Assuming the size of the feature map is $128^2$ and the attention scores are computed from $32^2$ patches, then the patch sizes should be greater or equal to $(128/32)^2=4^2$ so that all pixels can be covered. If the patch sizes are greater than $4\times4$, then certain pixels are overlapped, which is fine as the following layers of the network can learn to adapt. 

\noindent{\textbf{Multi-scale attention transfer and score sharing.}} In our framework, we apply attention transfer multiple times with the same set of attention scores (Figure~\ref{fig:method}). The sharing of attention scores leads to fewer parameters and better efficiency in terms of memory and speed.

\subsubsection{Residual Aggregation}
The target of Residual Aggregation is to calculate residuals for the hole region so that sharp details of the missing contents could be recovered. The residuals for the missing contents can be calculated by aggregating the weighted contextual residuals obtained from previous steps:

\begin{equation}
 R_j = \Sigma_{i=1}^N { s_{i,j} R_i}\\
\end{equation}

\noindent where $R$ is the residual image and $R_i$ is the $i^{th}$ patch extracted from contextual residual image outside the mask, and $R_j$ is $j^{th}$ patch to be filled inside the mask. The patch sizes are properly chosen to exactly cover all pixels without overlapping, to ensure the filled residuals being consistent with surrounding regions. Once the aggregated residual image is obtained, we add it to the up-sampled blurry image of the generator, and obtain a sharp result (Figure~\ref{fig:method}).

\subsection{Architecture of Generator}
\label{sect:network}
The network architecture of the generator is shown in Figure~\ref{fig:method}. We use a two-stage coarse-to-fine network architecture where the coarse network hallucinates rough missing contents, and the refine network predicts finer results. The generator takes an image and a binary mask indicating the hole regions as input and predicts a completed image. The input and output sizes are expected to be $512\times512$. In order to enlarge the perceptive fields and reduce computation, inputs are down-sampled to 256$\times$256 before convolution in the coarse network, different from the refine network who operates on $512\times512$. The prediction of the coarse network is naively blended with the input image by replacing the hole region of the latter with that of the former as the input to the refine network. Refine network computes contextual attention scores with a high-level feature map and performs attention transfer on multiple lower-level feature maps, thus distant contextual information can be borrowed at multiple abstraction levels. We also adopt dilated convolutions~\cite{iizuka2017globally} in both coarse and refine networks to further expand the size of the receptive fields. To improve the computational efficiency, our inpainting network is designed in a slim and deep fashion, and the LWGC is applied for all layers of the generator. Other implementation considerations include: (1) using `same' padding and ELUs~\cite{clevert2015fast} as activation for all convolution layers, (2) removing batch normalization layer as they deteriorate color coherency~\cite{iizuka2017globally}. 

\subsection{Light Weight Gated Convolution} 
Gated Convolutions (GC)~\cite{yu2018free} leverages the art of irregular hole inpainting. However, GC almost doubles the number of parameters and processing time in comparison to vanilla convolution. In our network, we proposed three modified versions of GC called Light Weight Gated Convolutions (LWGC), which reduces the number of parameters and processing time while maintaining the effectiveness. The output of the original GC can be expressed as:

\begin{equation}
\begin{split}
G = \textrm{conv}(W_g, I) \\
F = \textrm{conv}(W_f, I) \\
O = \sigma(G) \odot \psi(F) \\
\end{split}
\label{eq:gated}
\end{equation}

\noindent where $\sigma$ is Sigmoid function thus the output values are within $[0,1]$.~$\psi$ is an activation function which are set to ELU in our experiments.~$w_g$ and $w_f$ are two different set of convolutional filters, which are used to compute the gates and features respectively. GC enables the network to learn a dynamic feature selection mechanism. The three variations of LWGC that we propose are named as: depth-separable LWGC (LWGC$^{ds}$), pixelwise LWGC (LWGC$^{pw}$), and single-channel LWGC (LWGC$^{sc}$). They differ by the computation of the gate branch, G:

\begin{equation}
G = \textrm{conv}^{depth-separable}(W_g, I) 
\label{eq:gated_ds}
\end{equation}

\begin{equation}
G = \textrm{conv}^{pixelwise}(W_g, I) 
\label{eq:gated_1by1}
\end{equation}

\begin{equation}
G = \textrm{conv}(W_g, I), \textrm{ G is single-channel}
\label{eq:gated_d1}
\end{equation}

The depth-separable LWGC employs a depth-wise convolution followed by a $1\times1$ convolution to compute gates. The pixelwise LWGC uses a pixelwise or $1\times1$ convolution to compute the gates. The single-channel LWGC outputs a single-channel mask that is broadcast to all feature channels during multiplication. The single-channel mask is similar to partial convolution, whereas the mask of partial convolutions is hard-wired and untrainable, and generates a binary mask instead of a soft mask. Given that the height ($H_k$) and width ($W_k$) of kernels, and numbers of input channels ($C_{in}$) and output channels ($C_{out}$), we compare the number of parameters needed to calculate gates in Table \ref{table:gc-conv}. We used the single-channel LWGC for all layers of the coarse network and depth-separable or pixelwise LWGC for all layers of the refine network, which has been proved to be equally effective as regular GC but more efficient (Section \ref{sect:lwgc}).

\begin{table}[t]
	\centering
	\caption{Number of parameters needed to compute gates}
	\label{table:gc-conv}
	\small\addtolength{\tabcolsep}{-2pt}
	\begin{tabular}{l||c|c}
	       Method&   Parameters Calculation &  $H_k, W_k=3$  \\
	       &    &  $C_{in}, C_{out}=32$  \\
	      \hline
	      \hline
	     GC~\cite{yu2018free}          & $H_k\times W_k\times C_{in}\times C_{out}$  & 9216 \\
	LWGC$^{ds}$    & $H_k\times W_k\times C_{in} + C_{in}\times C_{out}$  & 1312 \\
	LWGC$^{pw}$  & $C_{in}\times C_{out}$  &  1024 \\
	LWGC$^{sc}$    & $H_k\times W_k\times C_{in} \times 1$  & 288
	\end{tabular}
\end{table}

\subsection{Training of the network}
\subsubsection{Training Losses}
Without the degradation of performance, we also significantly simplify the training objectives as two terms: the adversarial loss and the reconstruction loss. We use the WGAN-GP loss as our adversarial loss~\cite{gulrajani2017improved}, which enforces global consistency in the second-stage refinement network. The discriminator and generator are alternatively trained with the losses defined in Equation~\ref{eq:dloss} and Equation~\ref{eq:wgan_gp_gloss}:

\begin{equation}
\begin{split}
L_{d} = \mathbb{E}_{\tilde{\mathbf{x}} \in \mathbb{P}_g} [D(\tilde{\mathbf{x}})] - \mathbb{E}_{\mathbf{x} \in \mathbb{P}_r} [D(\mathbf{x})] + \\
\sigma \mathbb{E}_{\hat{\mathbf{x}} \in \mathbb{P_{\hat{\mathbf{x}}}}} [\| \bigtriangledown_{\hat{\mathbf{x}}} D(\hat{\mathbf{x}})\|_2 - 1]^2
\label{eq:dloss}
\end{split}
\end{equation}

\noindent where $D(.)$ is the discriminator output and $G(.)$ is the generator output. $\mathbf{x}$, $\mathbf{\tilde{x}}$, $\mathbf{\hat{x}}$, are real images, generated images, and interpolations between them, respectively. $\mathbb{P}_g$, $\mathbb{P}_r$, $\mathbb{P}_{\hat{\mathbf{x}}}$ are the corresponding distributions of them respectively.

\begin{equation}
L_{adv} = - \mathbb{E}_{\tilde{\mathbf{x}} \in \mathbb{P}_g} [D(\tilde{\mathbf{x}})] 
\label{eq:wgan_gp_gloss}
\end{equation}

We also add the $L1$ loss to force the consistency of the prediction with the original image. In contrast to~\cite{yu2018generative}, we avoid the computationally expensive spatially-discounted reconstruction loss. For simplicity, we just assign a smaller constant weight for the reconstruction loss of all in-hole pixels. The reconstruction loss is thus written as:

\small
\begin{equation}
L_{in-hole} = | G(\mathbf{x}, \mathbf{m})- \mathbf{x} | \odot \mathbf{m}  \\
\label{eq:loss_inhole}
\end{equation}

\begin{equation}
L_{context} =  | G(\mathbf{x}, \mathbf{m})- \mathbf{x} | \odot ( 1- \mathbf{m}) 
\label{eq:loss_context}
\end{equation}

\begin{equation}
L_{rec} =  \alpha_1 L_{in-hole} + \alpha_2 L_{context} 
\label{eq:reconstruction_loss}
\end{equation}

\normalsize
\noindent where $\alpha_1$ and $\alpha_2$ are coefficients for the in-hole term and contextual term ($\alpha_1=1$, and $\alpha_2=1.2$). The coarse network is trained with the reconstruction loss explicitly, while the refinement network is trained with a weighted sum of the reconstruction and GAN losses. The coarse network and refine network are trained simultaneously with merged losses as shown in Equation~\ref{eq:gloss}.
\begin{equation}
L_g = L_{rec} + \beta L_{adv}  
\label{eq:gloss}
\end{equation}

\noindent where $\beta$ is the coefficient for adversarial loss ($\beta=10^{-4}$).

\begin{figure*}[t]
\begin{center}
  \includegraphics[width=\linewidth]{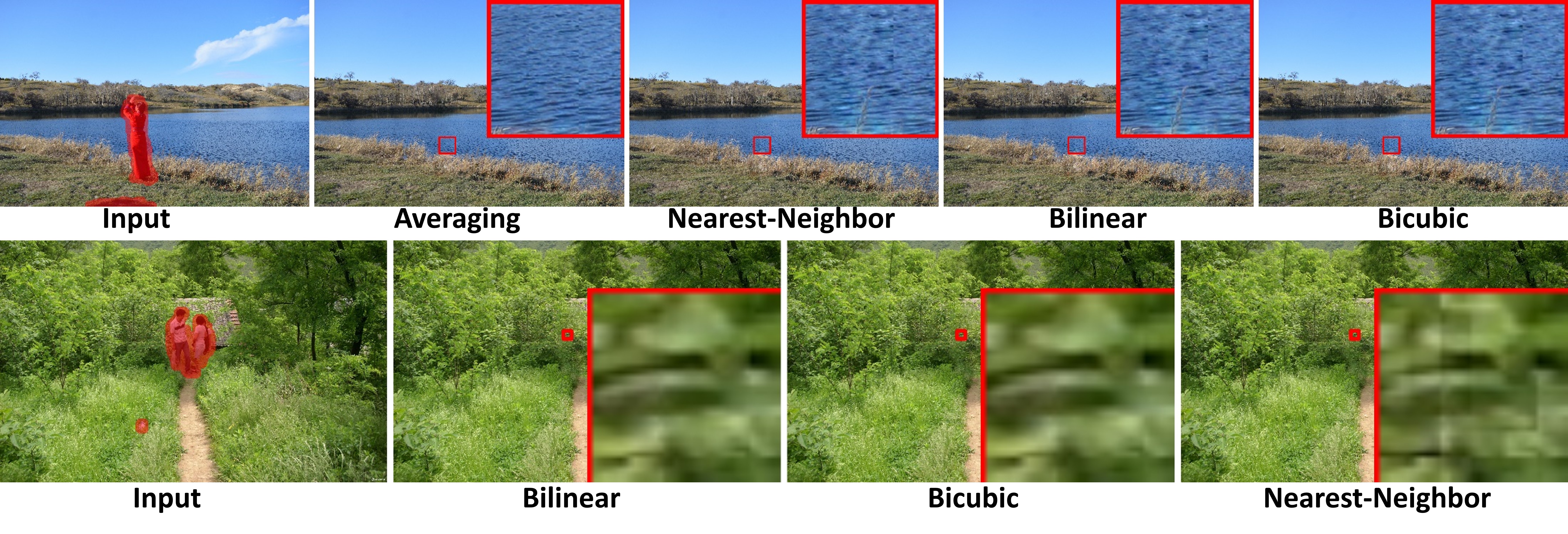} 
\end{center}
   \caption{Comparing down-sampling and up-sampling operators: (top) using Bilinear up-sampling along with Averaging down-sampling generates more coherent textures with the surroundings. (bottom) using the Averaging down-sampling along with Nearest Neighbor produces tiling artifacts while Bilinear and Bicubic up-sampling perform equally well.}
\label{fig:amblation-downup}
\end{figure*}

\begin{figure*}[t]
\begin{center}
  \includegraphics[width=\linewidth]{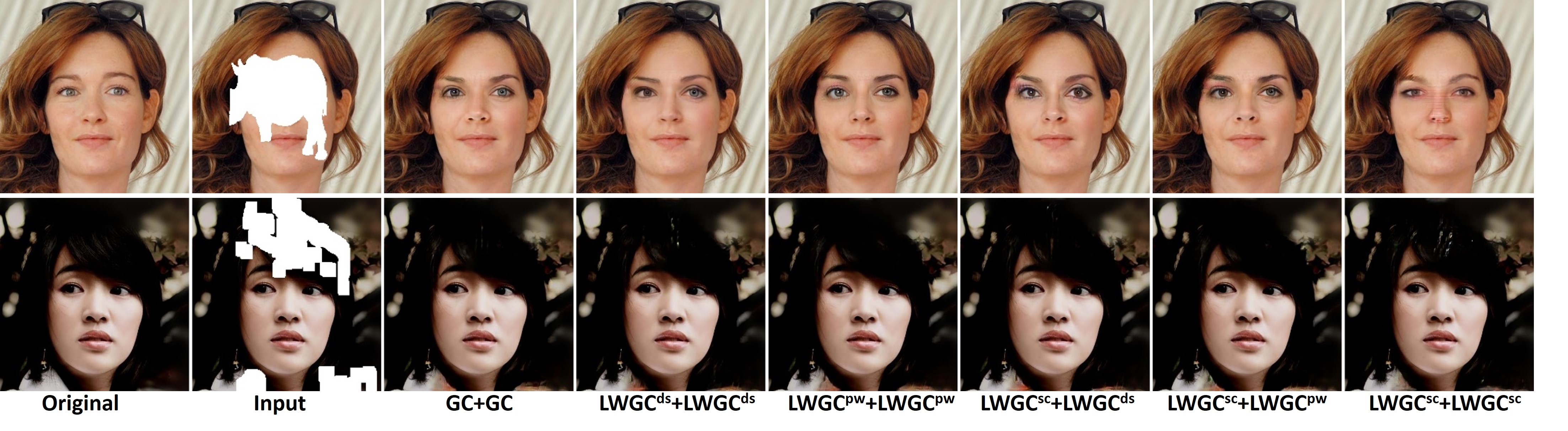} 
\end{center}
   \caption{Comparisons of different Gated Convolutions configurations. For example, the notation of LWGC$^{sc}$+LWGC$^{ds}$ means: the coarse network uses single-channel LWGC and the refine network uses depth-separable LWGC.}
\label{fig:amblation-gc}
\end{figure*}

\subsubsection{Random Mask Generation}
To diversify the inpainting masks, we use two methods to generate irregular masks on-the-fly during training. The first one is~\cite{liu2018image}, which simulates tears, scratches, spots or manual erasing with brushes. The second approach generates masks by randomly manipulating the real object shape templates, accounting for the object removal scenario. These shape templates are obtained from object segmentation masks and including a wide range of categories such as single, multiple or crowded objects. We also randomly rotate, flip and scale the templates with a random scale ratio. In practice, the aforementioned two methods can be combined or separated, depending on specific needs.

\subsubsection{Training Procedure}
During training, color values of all images are linearly scaled to $[-1,1]$ in all experiments, and the mask uses 1 to indicate the hole region and 0 to indicate background. The masked image is constructed as $\mathbf{x} \odot (1 - \mathbf{m})$, where $\mathbf{x}$ is the input image and $\mathbf{m}$ is the binary mask, and $\odot$ represents dot product. Inpainting generator G takes concatenation of the masked image and mask as input, and predicts $\mathbf{y} = G(\mathbf{x}, \mathbf{m})$ of the same size as the input image. The entire training procedure is illustrated in Algorithm~\ref{alg:procedure}.

\begin{algorithm}
	\small
	\caption{Training of our proposed network}
	\label{alg:procedure}
	\SetAlgoLined
	initialization; \\
	\While{G has not converged}{
		\For{i = 1,...,5}{
			Sampling batch images $\mathbf{x}$ from training data;\\
			Generating random masks $\mathbf{m}$ for $\mathbf{x}$;\\
			Getting inpainted $\mathbf{y} \leftarrow G(\mathbf{x}, \mathbf{m})$;\\
			Pasting back $\tilde{\mathbf{x}} \leftarrow \mathbf{y}  \odot \mathbf{m} + \mathbf{x}\odot (1-\mathbf{m})$;\\
			Sampling a random number $\alpha \in U[0,1]$;\\
			Getting interpolation $\mathbf{\hat{x}} \leftarrow (1-\alpha) \mathbf{x} + \alpha \tilde{\mathbf{x}}$;\\
			Updating the discriminator D with loss $L_d$;\\
		}
		Sampling batch images $\mathbf{x}$ from training data;\\
		Generating random masks $\mathbf{m}$ for $\mathbf{x}$;\\
		Getting Inpainted $\mathbf{y} \leftarrow G(\mathbf{x}, \mathbf{m})$;\\
		Pasting back $\tilde{\mathbf{x}} \leftarrow \mathbf{y}  \odot \mathbf{m} + \mathbf{x}\odot (1-\mathbf{m})$;\\
		Updating generator G with loss $L_g$;\\
	}
\end{algorithm}


\section{Experimental Results}
We evaluate the proposed method on three datasets including Places2~\cite{zhou2017places}, CelebA-HQ~\cite{karras2017progressive}, and DIV2K~\cite{Timofte_2018_CVPR_Workshops}. Our model is trained on two NVIDIA 1080 Ti GPUs with images of resolution $512\times512$ with batch size of 8. For DIV2K and CelebA-HQ, images are down-sampled to $512\times512$. For Places2, images are randomly cropped to $512\times512$. After training, we test the models on images of various resolutions of 512 to 8K on a GPU. The final model has a total of 2.7M parameters and implemented on TensorFlow v1.13 with CUDNN v7.6 and CUDA v10.0. 

\begin{figure*}[t]
	\begin{center}
		\includegraphics[width=\linewidth]{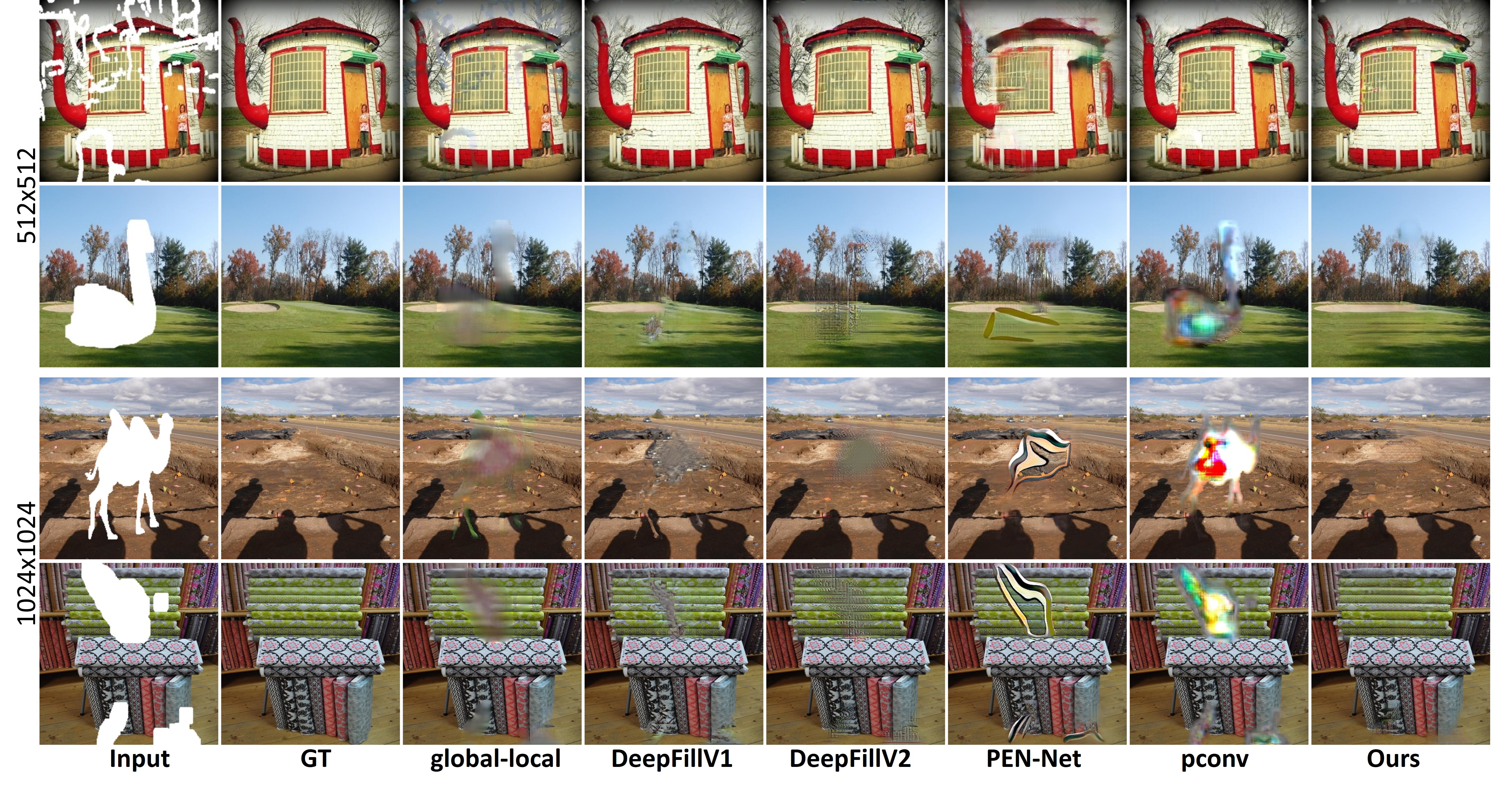} 
	\end{center}
	\caption{Qualitative comparisons using $512\times512$ (top) and $1024\times1024$ (bottom) images from Places2 validation dataset.}
	\label{fig:comparison1}
\end{figure*}

\begin{table}
	\centering
	\caption{Quantitative evaluation results on Places2 validation set. Note that certain models cause Out-Of-Memory (OOM) error when tested on 2K or 4K images, thus the corresponding cells are left empty.}
	\label{table:quant}
	\resizebox{\textwidth}{!}{
		\begin{tabular}{c||ccccc|ccccc|ccccc|ccccc}
			Image Size & \multicolumn{5}{|c}{$512\times512$} &  \multicolumn{5}{|c}{$1024\times1024$}  & \multicolumn{5}{|c}{$2048\times2048$}  & \multicolumn{5}{|c}{$4096\times4096$}  \\
			\hline
			Metrics & L1  &  MS-SSIM & FID & IS & Time & L1  &  MS-SSIM & FID & IS & Time & L1  &  MS-SSIM & FID & IS & Time& L1  &  MS-SSIM & FID & IS & Time\\
			\hline
			\hline
			DeepFillV1\cite{yu2018generative}  &  6.733  &  0.8442   &  7.541  & 17.56 & 62ms &   7.270  &  0.8424   &  10.21  & 17.69 & 663ms & -- &--& -- &--& -- &-- &--& -- &--&--\\
			DeepFillV2\cite{yu2018free}  &  6.050 &  \textbf{0.8848}   &  4.939  &  \textbf{18.20} & 78ms &  6.942 &  0.8784   &  8.347  &  17.04 & 696ms & -- &--& -- &--& -- &-- &--& -- &--&--\\
			PEN-Net\cite{zeng2019learning}   & 9.732  & 0.8280   & 14.13 & 14.51  & 35ms & 10.42  & 0.8128   & 19.36 & 12.51  & 289ms  & -- &--& -- &--& -- &-- & -- &--& -- &--\\
			PartialConv\cite{liu2018image} & 8.197 &    0.8399 &   29.32   & 13.13 &35ms & 11.19 &    0.8381 &   32.20   & 13.53 & 110ms  & 16.19 &    0.8373 &   41.23   & 11.17 & 920ms & -- &--& -- &--& --  \\
			Global-local\cite{iizuka2017globally} & 8.617  &  0.8469  &  21.27  &   13.48  &53ms & 9.232  &  0.8392  &  26.23  &   13.05  & 219ms & 9.308  &  0.8347  &  27.09  &   12.61  & 219ms & -- &--& -- &--& -- \\
			\textbf{Ours}  &  \textbf{5.439}  &   0.8840   & \textbf{4.898}  & 17.72 & \textbf{25ms}  &  \textbf{5.439}  &   \textbf{0.8840}   & \textbf{4.899}  & \textbf{17.72} & \textbf{31ms} &  \textbf{5.492}  &   \textbf{0.8840}   & \textbf{4.893}  & \textbf{17.85} & \textbf{37ms}  &  \textbf{5.503}  &   \textbf{0.8840}   & \textbf{4.895}  & \textbf{17.81} & \textbf{87.3ms} 
	\end{tabular}}
\end{table}

\subsection{Analysis of CRA Design}\label{sect:cra}
As shown in Figure~\ref{fig:method}, the CRA mechanism involves one down-sampling and two up-sampling operations outside of the generator. Choosing different methods for down-sampling and up-sampling may affect the final results. To explore this, we experimented with four down-sampling methods: Nearest-Neighbor, Bilinear, Bicubic, and Averaging. Averaging evenly splits the input into target patches and average all pixels in each patch to obtain a 512$\times$512 image. We also explored three up-sampling methods including Nearest-Neighbor, Bilinear or Bicubic. Note that the two up-sampling operations must be consistent, so we do not consider inconsistent combinations. Experimental results on an HD dataset indicate that Averaging performs the best for down-sampling and Bilinear or Bicubic performs equally well for up-sampling (Figure~\ref{fig:amblation-downup}). For simplicity, we use Averaging down-sampling and Bilinear up-sampling.

\subsection{Analysis of Light Weight Gated Convolutions}\label{sect:lwgc}
We propose three types of LWGC, which are faster than the original GC. We experimented how they affect inpainting quality and efficiency on the CelebA-HQ dataset to explore the influence of LWGC on the results, by replacing the original GCs with LWGCs for the coarse/refine networks. As shown in Figure~\ref{fig:amblation-gc}, the LWGC$^{sc}$+LWGC$^{sc}$ configuration brings visible artifacts while the other five configurations perform equally well in terms of quality. Considering LWGC$^{sc}$+LWGC$^{pw}$ requires fewer parameters than the other four, we adopt the LWGC$^{sc}$+LWGC$^{pw}$ configuration in the generator.

\subsection{Comparisons With Learning-based Methods}
We compared our method with other state-of-the-art learning-based inpainting methods including Global-local GAN~\cite{iizuka2017globally}, DeepFillV1~\cite{yu2018generative}, DeepFillV2~\cite{yu2018free}, PEN-Net~\cite{zeng2019learning} and Partial Convolution~\cite{liu2018image}. To make fair comparisons, we attempted to use the same settings for all experiments, though not fully guaranteed. The official pre-trained DeepFillV1~\cite{yu2018generative} model was trained for 100M iterations with the batch size of 16 and the global-local GAN~\cite{iizuka2017globally} was trained for 300K iterations with the batch size of 24. Both of them were trained on 256$\times$256 images with rectangular holes of maximum size 128$\times$128. All the other models were trained for 300K iterations with the batch size of 8 on 512$\times$512 images with irregular holes up to 25\% area of the whole image. The original DeepFillV2~\cite{yu2018free} model attached a sketch channel to the input to facilitate image manipulation, we simply removed the sketch channel and re-trained the model. For all these methods, no specific post-processing steps are performed other than pasting the filled contents back to the original image.

\begin{figure*}[h]
	\begin{center}
		\includegraphics[width=\linewidth]{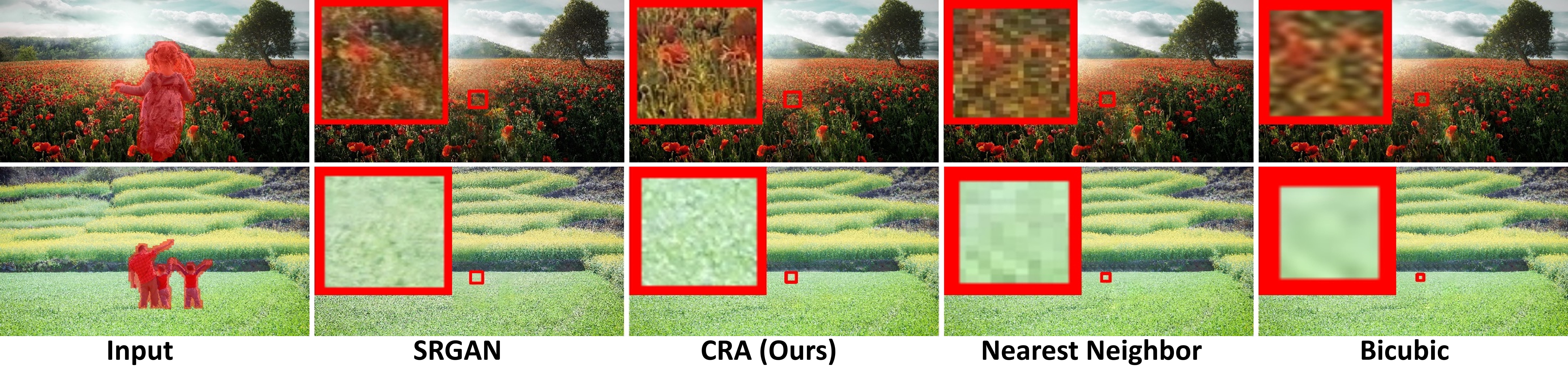} 
	\end{center}
	\caption{Comparisons of different super-resolution methods: the red squares area are zoomed-in for more details.}
	\label{fig:hd}
\end{figure*}


\begin{figure*}[h]
\begin{center}
  \includegraphics[width=\linewidth]{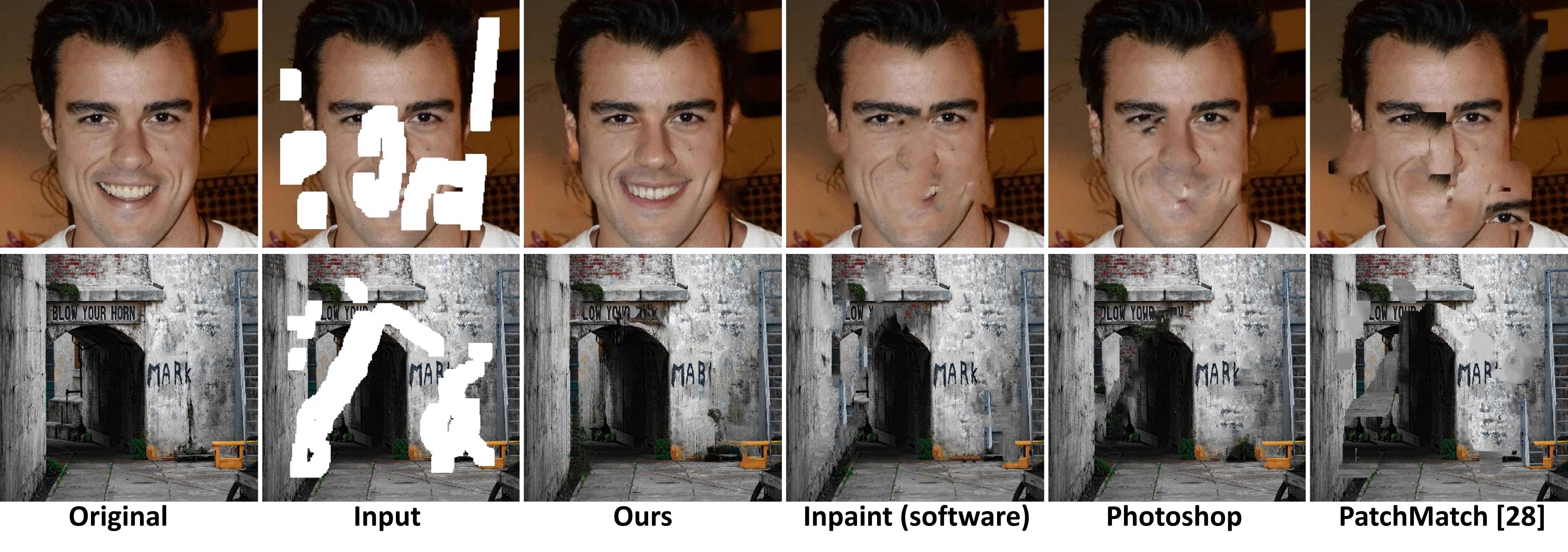} 
\end{center}
   \caption{Comparisons of our method with Inpaint (software), Photoshop content-aware fill and an open-source PatchMatch implementation~\cite{Hsieh_patchmatch2019}. The masks for Photoshop and Inpaint are manually drawn, thus not guaranteed to be the same. }
\label{fig:hd4}
\end{figure*}

\noindent\textbf{Qualitative comparisons} Figure~\ref{fig:comparison1} shows our model performs equally good or slightly better than previous methods on $512\times512$ images. Partial convolution~\cite{liu2018image} and global-local GAN~\cite{iizuka2017globally} performs well when the mask is small and narrow but exert severe artifacts when the hole size becomes bigger. Global-local GAN~\cite{iizuka2017globally} is problematic in maintaining the color consistency of filled contents with surroundings. DeepFillV1~\cite{yu2018generative} generates plausible results, but occasionally the artifacts inside the hole region are visible, implying its vulnerability to irregular masks. DeepFillV2~\cite{yu2018free} generates incoherent textures when the hole size goes up. Nevertheless, when tested on larger images with bigger hole sizes, our model performs consistently good while the inpainting results of other methods deteriorate dramatically (e.g. severe hole-shaped artifacts in Figure~\ref{fig:comparison1}).

\noindent\textbf{Quantitative comparisons} Table~\ref{table:quant} reports our quantitative evaluation results in terms of mean $L1$ error, MS-SSIM~\cite{wang2003multiscale}, Inception Score (IS)~\cite{salimans2016improved}, and Frechet Inception Distance (FID)~\cite{heusel2017gans}. It also shows the average inference time per image on a NVIDIA GTX 1080 Ti GPU. These metrics are calculated over all 36,500 images of the Places2 validation set. Each image is assigned a randomly generated irregular mask. To examine the performance on various image sizes, we linearly scale images and masks to various dimensions. Table~\ref{table:quant} shows that our proposed model achieves the lowest $L1$ loss and FID on 512$\times$512 images. When the input images are greater than or equal to 1024$\times$1024, our proposed model achieves the best results on all metrics. In addition, the proposed approach significantly outperforms other learning-based methods in terms of speed. In specific, it is 28.6\% faster for 512$\times$512 images, 3.5 times faster for 1024$\times$1024 images, and 5.9 times faster for 2048$\times$2048 images than the second-fastest method. Furthermore, the proposed model can inpaint 4096$\times$4096 images in 87.3 milliseconds which is intractable with other learning-based methods due to limits of GPU memory.

\subsection{Comparisons of CRA with Super-resolution} 
Figure~\ref{fig:hd}, compares the high-resolution results of our CRA with those obtained via various super-resolution techniques. After obtaining a 512$\times$512 inpainted result, we up-sample the output to the original size using different up-sampling methods including SRGAN~\cite{ledig2017photo}, Nearest Neighbor, and Bicubic, then, we paste the filled contents to the original image. SRGAN~\cite{ledig2017photo} is a learning-based method that can perform 4$\times$ super-resolution and the official pre-trained model was trained on DIV2K. We can observe from that the hole region generated by CRA is generally sharper and visually more consistent with surrounding areas.

\subsection{Comparisons With Traditional Methods} 
Moreover, we compare our method with two commercial products based on PatchMatch~\cite{barnes2009patchmatch} (Photoshop, Inpaint) and an open-source PatchMatch implementation for image inpainting~\cite{Hsieh_patchmatch2019} (Figure~\ref{fig:hd4}). We discover that PatchMatch-based methods are able to generate clear textures but with distorted structures incoherent with surrounding regions.

\section{Conclusion}
We presented a novel Contextual Residual Aggregated technique that enables more efficient and high-quality inpainting of ultra high-resolution images. Unlike other data-driven methods, the increase of resolutions and hole size does not deteriorate the inpainting quality and does not considerably increase the processing time in our framework. When tested on high-resolution images between 1K and 2K, our model is extremely efficient where it is 3x$\sim$6x faster than the state-of-the-art on images of the same size. Also, it achieves better quality by reducing FID by 82\% compared to the state-of-the-art. We also compared our method with commercial products that showed significant superiority in certain scenarios. So far, our method is the only learning-based technique that can enable end-to-end inpainting on the ultra-high-resolution image (4K$\sim$8K). In the future, we will explore similar mechanisms for other tasks such as image expansion, video inpainting and image blending.

\section*{Acknowledgment}
We want to acknowledge Peng Deng, Shao Hua Chen, Xinjiang Sun, Chunhua Tian and other colleagues in Huawei Technologies for their support in the project.

\bibliographystyle{splncs}
\bibliography{hifill}
\section*{Appendix}
\subsection*{Network Architectures}
In addition to Section 3.3 and Figure 2 in the main paper, we report more details of our network architectures. For simplicity, we denote them with
K (kernel size), S (stride size), C (channel number) and D (dilation rate). D is neglected when D=1.

\paragraph{Coarse Network:}
downsample(2$\times$) - K5S2C32 - K3S1C32 - K3S2C64 - K3S1C64 - K3S1C64 - K3S1C64 - K3S1C64 - K3S1C64 - K3S1C64 - K3S1C64D2 - K3S1C64D2 - K3S1C64D2 - K3S1C64D2 - K3S1C64D2 - K3S1C64D4 - K3S1C64D4 - K3S1C64D4 - K3S1C64D4 - K3S1C64D8 - K3S1C64D8 - K3S1C64 - K3S1C64 - K3S1C64 - upsample(2$\times$) - K3S1C32 - upsample(2$\times$) - K3S1C3 - clip - upsample(2$\times$)

\paragraph{Refine Network:}
K5S2C32 - K3S1C32[$P^{l=1}$] - K3S2C64 - K3S1C64[$P^{l=2}$] - K3S2C128 - K3S1C128 - K3S1C128 - K3S1C128D2 - K3S1C128D4 - K3S1C128D8 - K3S1C128D16[$P^{l=3}$] - concat - K3S1C128 - upsample(2$\times$) - K3S1C64 - K3S1C64 - concat - upsample(2$\times$) - K3S1C32 - K3S1C32 - concat - upsample(2$\times$) - K3S1C3 - clip

\paragraph{Attention Computing Branch:}
[$P^{l=3}$] - downsample (2$\times$) - [$P$] - ACM - ATM

\paragraph{Attention Transfer Branch ($P^{l=3}$):}
[$P^{l=3}$] - ATM - K3S1C128 - concat

\paragraph{Attention Transfer Branch ($P^{l=2}$):}
[$P^{l=2}$] - ATM - K3S1C64 - K3S1C64D2 - concat

\paragraph{Attention Transfer Branch ($P^{l=1}$):}
[$P^{l=1}$] - ATM - K3S1C32 - K3S1C32D2 - concat

\paragraph{Discriminator: }
K3S2C64 - K3S2C128 - K3S2C256 - K3S2C256 - K3S2C256 - K3S2C256 - fully connected to 1.

\subsection*{More Test Results on Places2}
\begin{figure*}[t]
	\begin{center}
		\includegraphics[width=\linewidth]{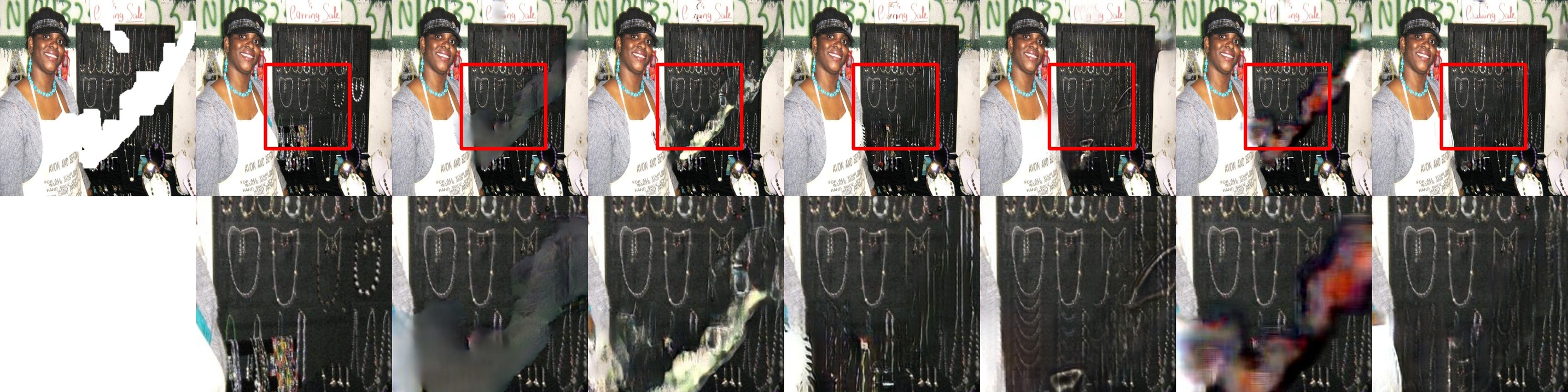}
		\includegraphics[width=\linewidth]{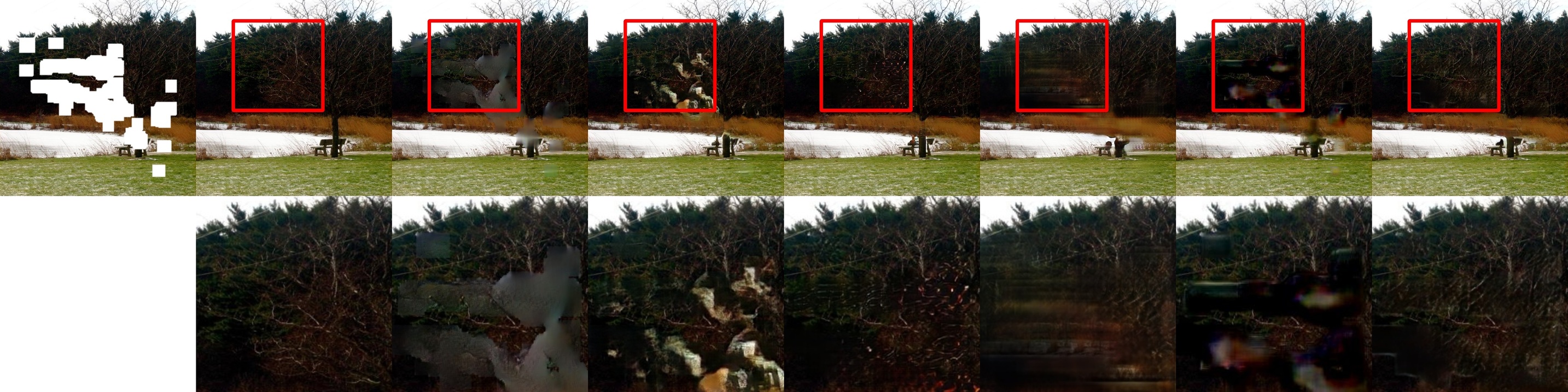}
		\includegraphics[width=\linewidth]{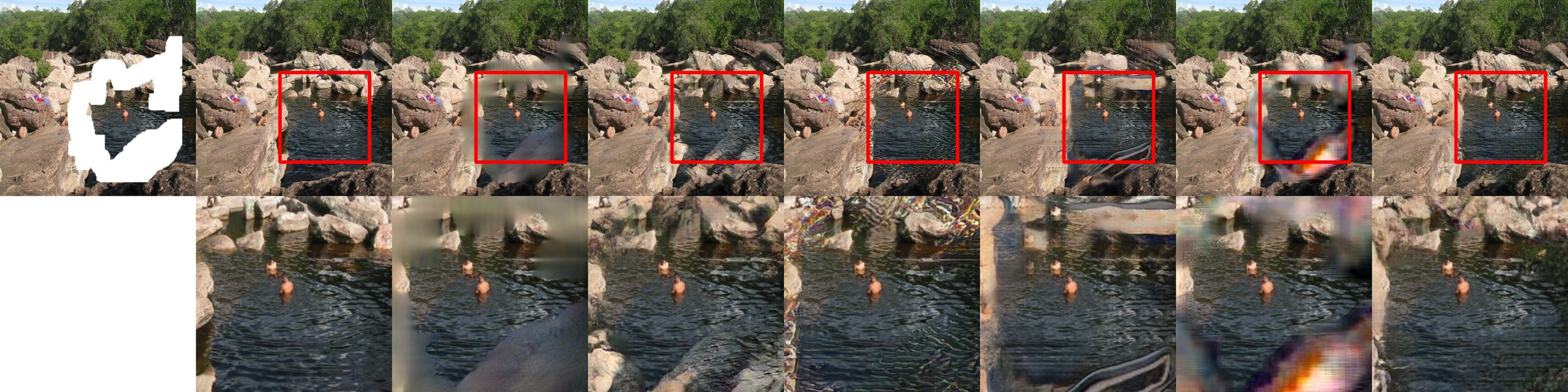}
		\includegraphics[width=\linewidth]{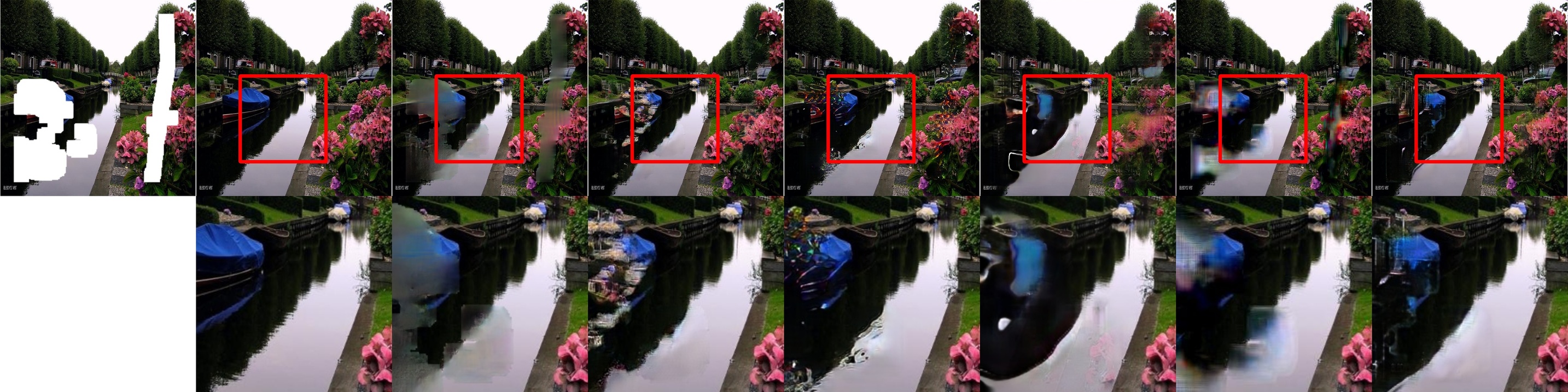}
		\subfloat[Input ]{\hspace{.125\linewidth}}
		\subfloat[GT]{\hspace{.125\linewidth}}
		\subfloat[global-local ]{\hspace{.125\linewidth}}
		\subfloat[DeepFillV1]{\hspace{.125\linewidth}}
		\subfloat[DeepFillV2 ]{\hspace{.125\linewidth}}
		\subfloat[PEN-Net]{\hspace{.125\linewidth}}
		\subfloat[pconv ]{\hspace{.125\linewidth}}
		\subfloat[Ours]{\hspace{.125\linewidth}}
		\caption{Test results on places2 validation datasets with input size of 512 $\times$ 512.}
		\label{fig:places512}
	\end{center}
\end{figure*}

\begin{figure*}[t]
	\begin{center}
		\includegraphics[width=\linewidth]{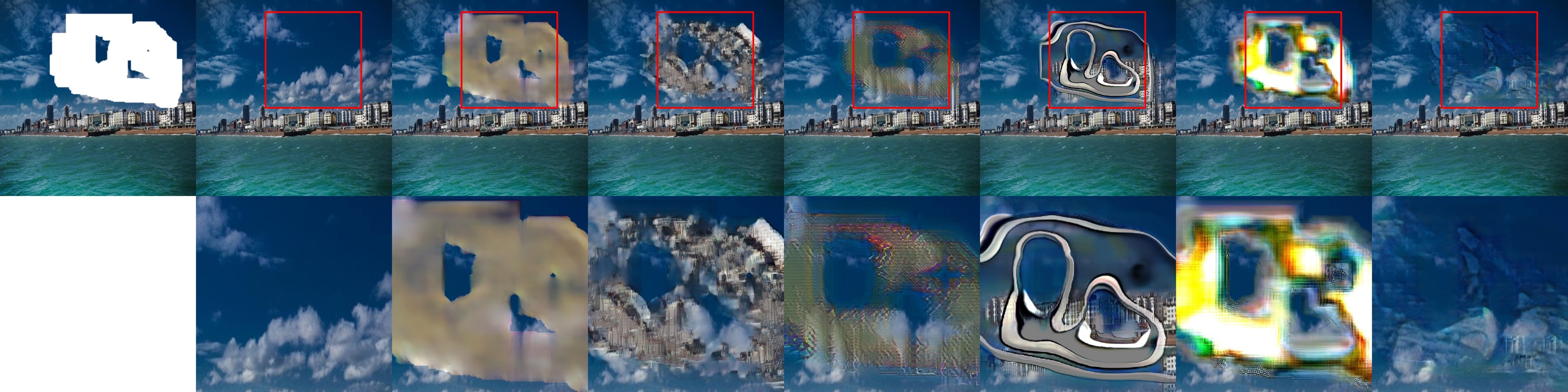}
		\includegraphics[width=\linewidth]{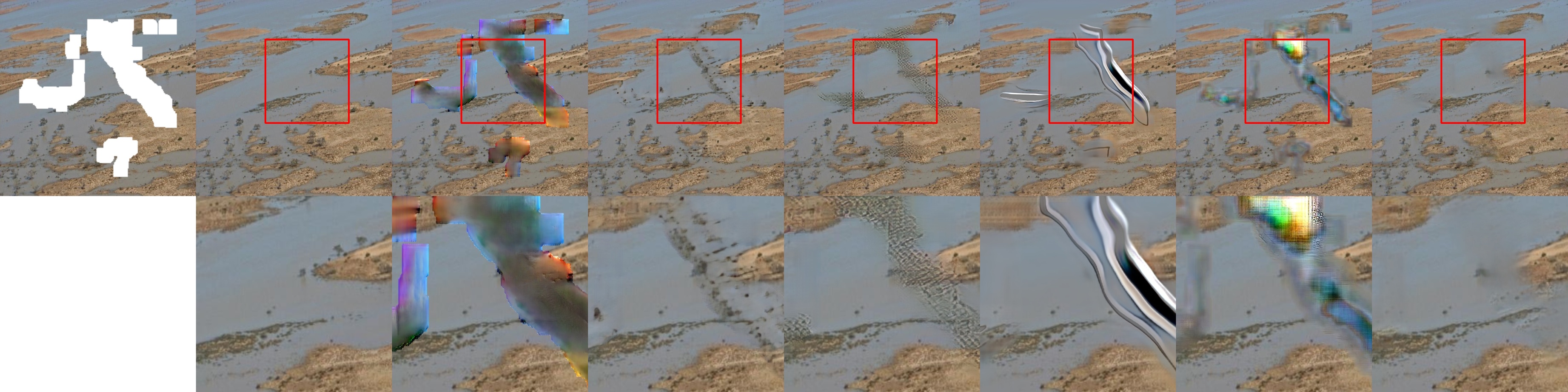}
		\includegraphics[width=\linewidth]{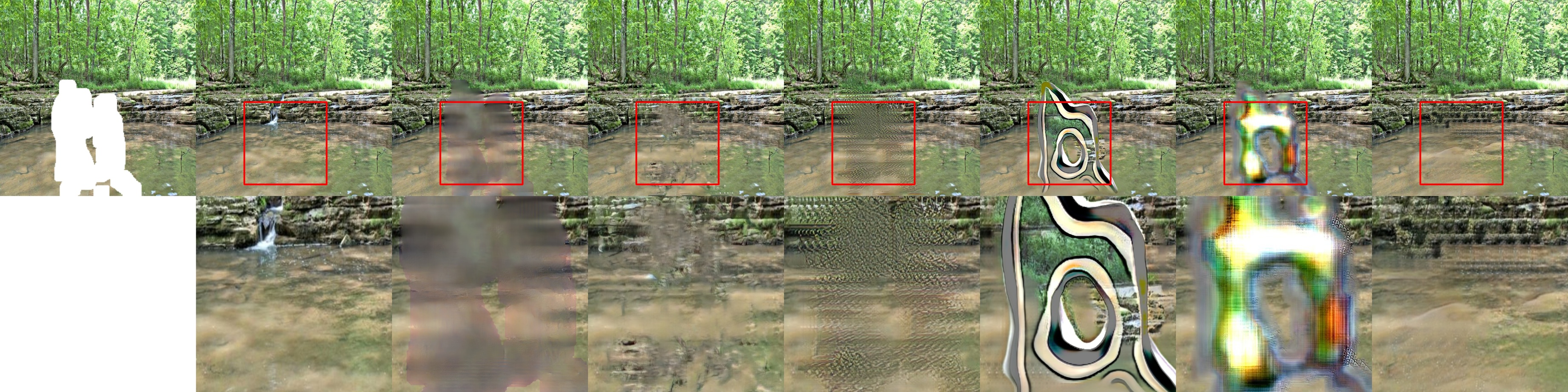}
		\includegraphics[width=\linewidth]{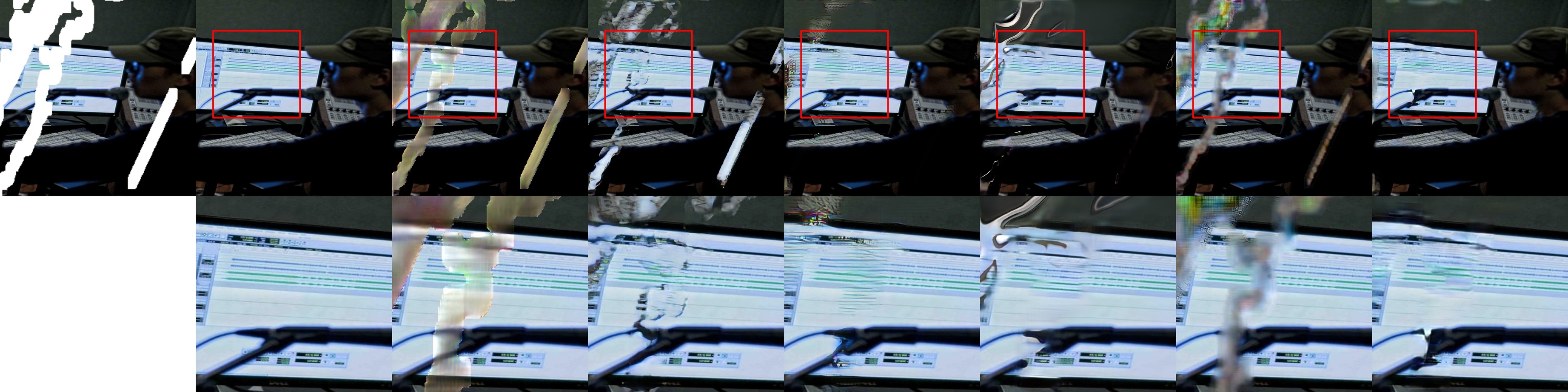}
		\subfloat[Input ]{\hspace{.125\linewidth}}
		\subfloat[GT]{\hspace{.125\linewidth}}
		\subfloat[global-local ]{\hspace{.125\linewidth}}
		\subfloat[DeepFillV1]{\hspace{.125\linewidth}}
		\subfloat[DeepFillV2 ]{\hspace{.125\linewidth}}
		\subfloat[PEN-Net]{\hspace{.125\linewidth}}
		\subfloat[pconv ]{\hspace{.125\linewidth}}
		\subfloat[Ours]{\hspace{.125\linewidth}}
		\caption{Test results on places2 validation datasets with input size of 1024 $\times$ 1024.}
		\label{fig:places512-1k}
	\end{center}
\end{figure*}

\begin{figure*}[t]
	\begin{center}
		\includegraphics[width=\linewidth]{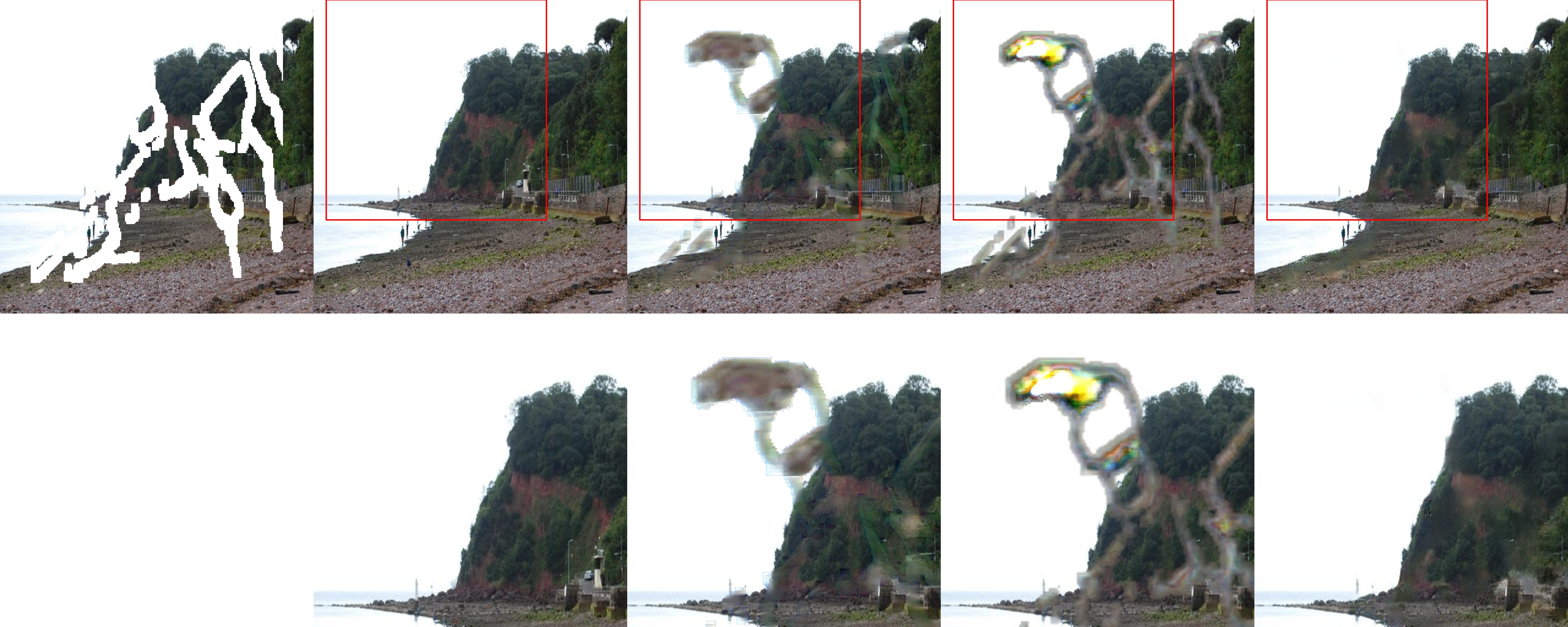}
		\includegraphics[width=\linewidth]{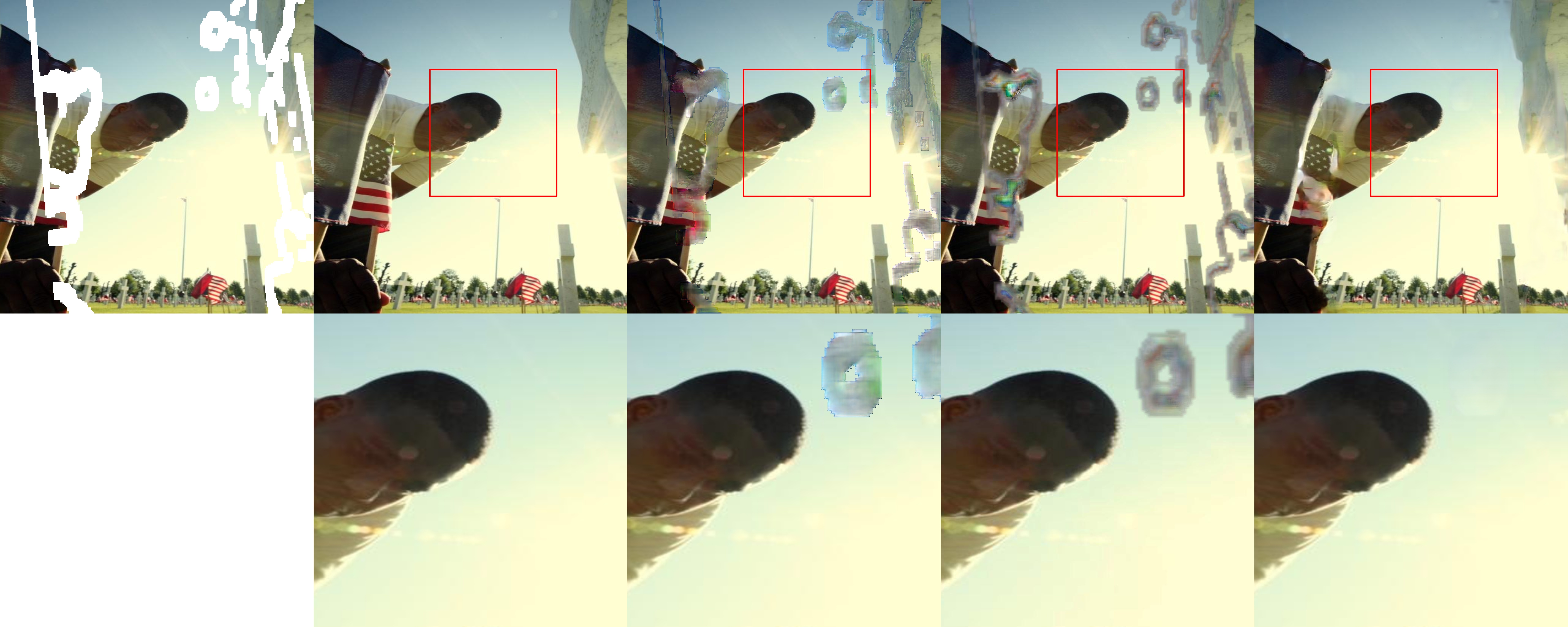}
		\includegraphics[width=\linewidth]{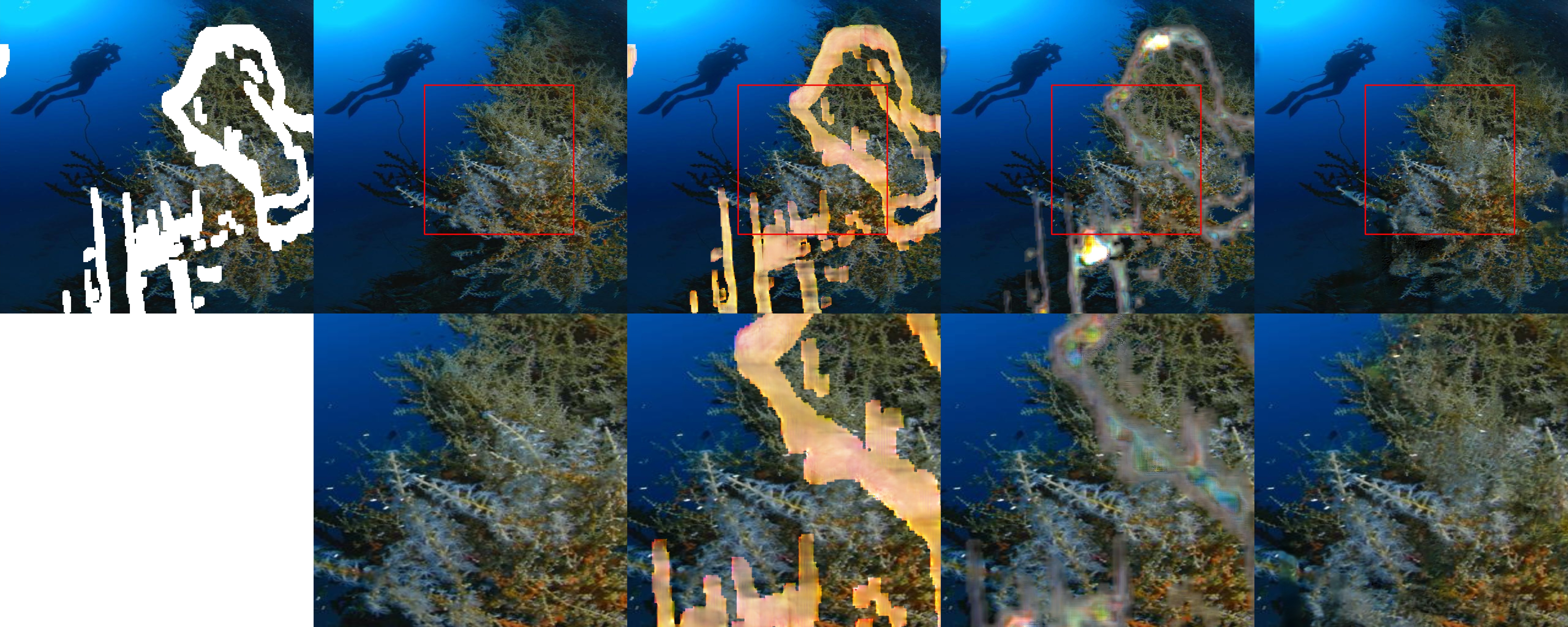}
		\subfloat[Input ]{\hspace{.2\linewidth}}
		\subfloat[GT]{\hspace{.2\linewidth}}
		\subfloat[global-local ]{\hspace{.2\linewidth}}
		\subfloat[pconv ]{\hspace{.2\linewidth}}
		\subfloat[Ours]{\hspace{.2\linewidth}}
		\caption{Test results on places2 validation datasets with input size of 2048 $\times$ 2048.}
		\label{fig:places512-2k}
	\end{center}
\end{figure*}

More test results on places2 are presented in Figures \ref{fig:places512}, \ref{fig:places512-1k}, \ref{fig:places512-2k}, with input size 512$\times$512, 1024$\times$1024, 2048$\times$2048 respectively. 

\subsection*{Sources of High-Resolution Images}
Sources of the HD images in the main paper that are crawled from the internet are presented in Table \ref{table:source}.

\begin{table}[t]
	\centering
	\caption{Sources of some HD images used for test}
	\label{table:source}
	\resizebox{\textwidth}{!}{
	\begin{tabular}{l|l}
		Figure ID \\  in the main paper & Image Source \\
		\hline
		{\tiny Figure 3 top}  & {\tiny\url{http://www.sohu.com/a/117062677_189010}}   \\
		{\tiny Figure 6 top}    & {\tiny\url{http://ow.ly/u8Wff} } \\
		{\tiny Figure 6 bottom}  & {\tiny\url{https://www.mafengwo.cn/yj/14103/s-0-0-0-0-1-0.html}}  \\
		{\tiny Figure 1 topright}  & {\tiny\url{https://www.champaignoutdoors.com/kilimanjaro}}  \\
		{\tiny Images in demo.pps}    & {\tiny\url{http://www.imecchina.com/news/1293274.html}}  \\
		& {\tiny\url{http://www.zdqx.com/wall/57962_6.html}}  \\
		& {\tiny\url{https://www.xuehua.us/2018/06/03/\%E5\%92\%8C\%E9\%AB\%98\%E5\%B0\%94\%E5\%A4}} \\
				& {\tiny\url{\%AB\%E5\%98\%89\%E6\%97\%85\%E4\%B8\%80\%E9\%81\%93-\%E6\%8E\%A2\%E5\%AF\%BB\%E4\%BB}}\\
						& {\tiny\url{\%99\%E6\%B9\%96\%E8\%BE\%B9\%E7\%9A\%84\%E6\%85\%A2\%E7\%94\%9F\%E6\%B4\%BB/zh-tw/}}\\
								& {\tiny\url{https://you.autohome.com.cn/details/68005/727cc0cec7214dd62e92d8f009e7adf9}}  \\
								& {\tiny\url{https://www.reyfoto.com/}}
								
	\end{tabular}}
\end{table}
						
\subsection*{Failure Examples \& Limitation}
\begin{figure*}[t]
\begin{center}
\includegraphics[width=.49\linewidth]{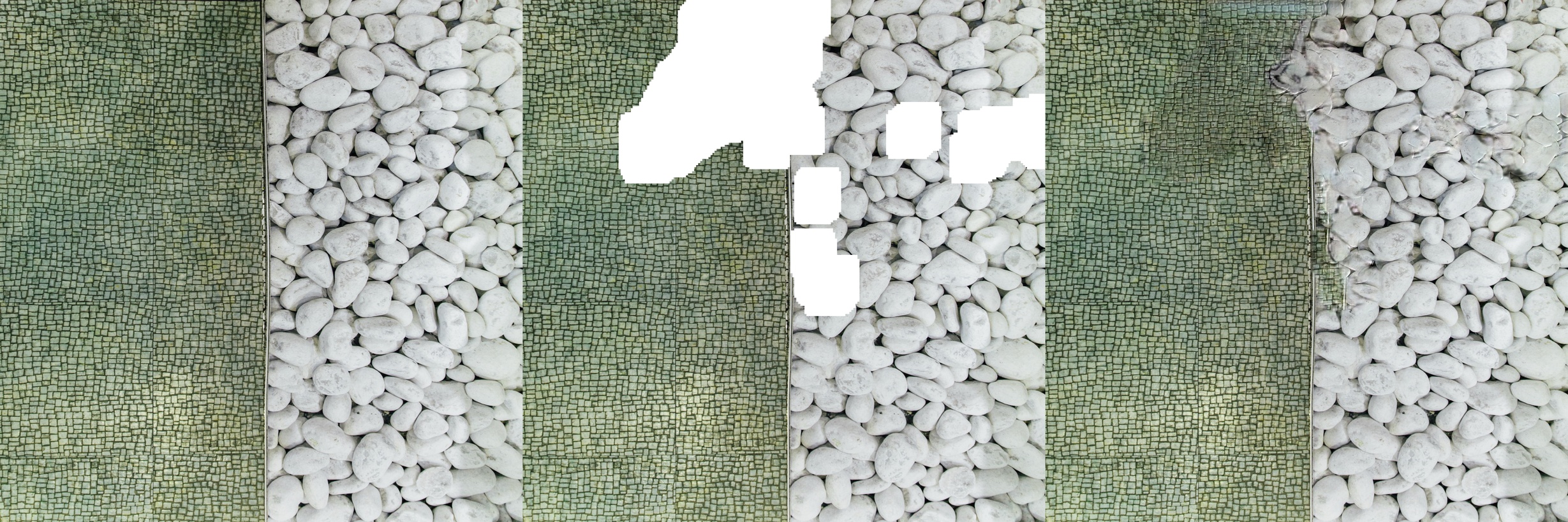}
\includegraphics[width=.49\linewidth]{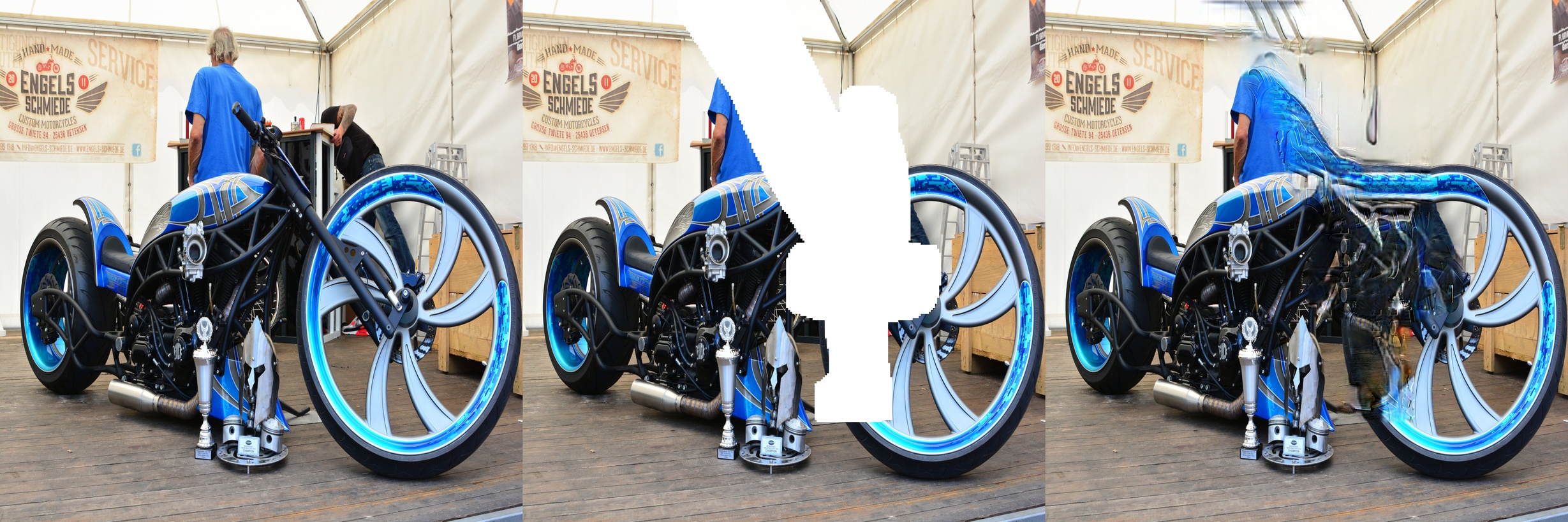}
\includegraphics[width=.49\linewidth]{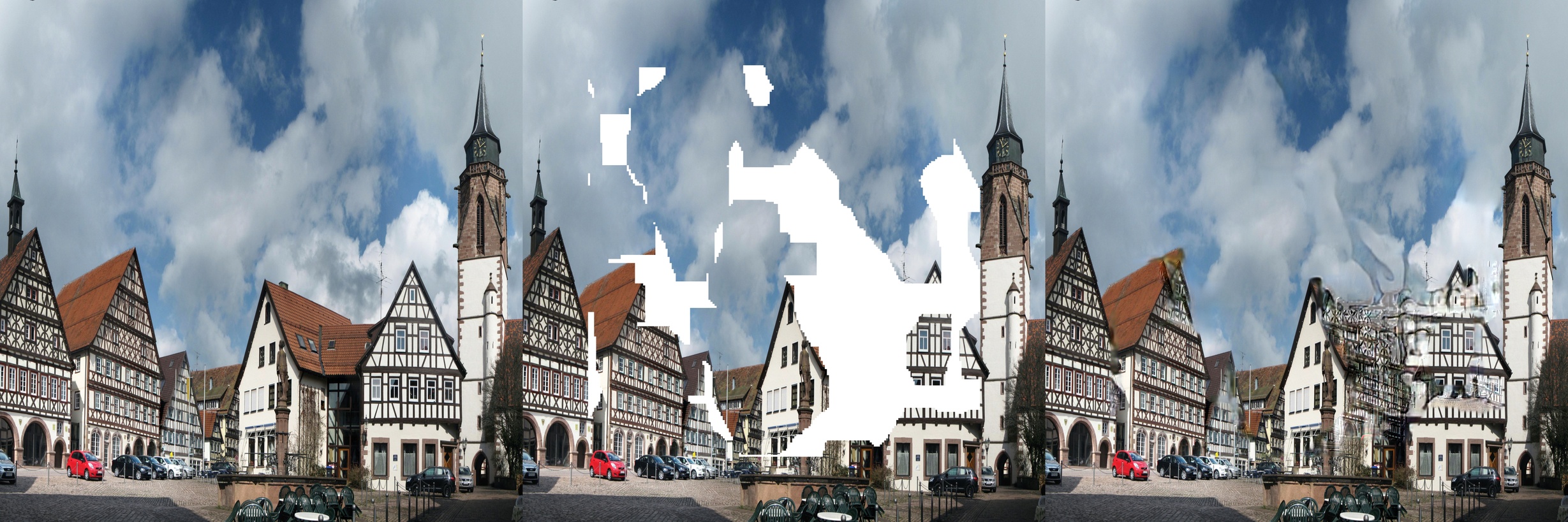}
\includegraphics[width=.49\linewidth]{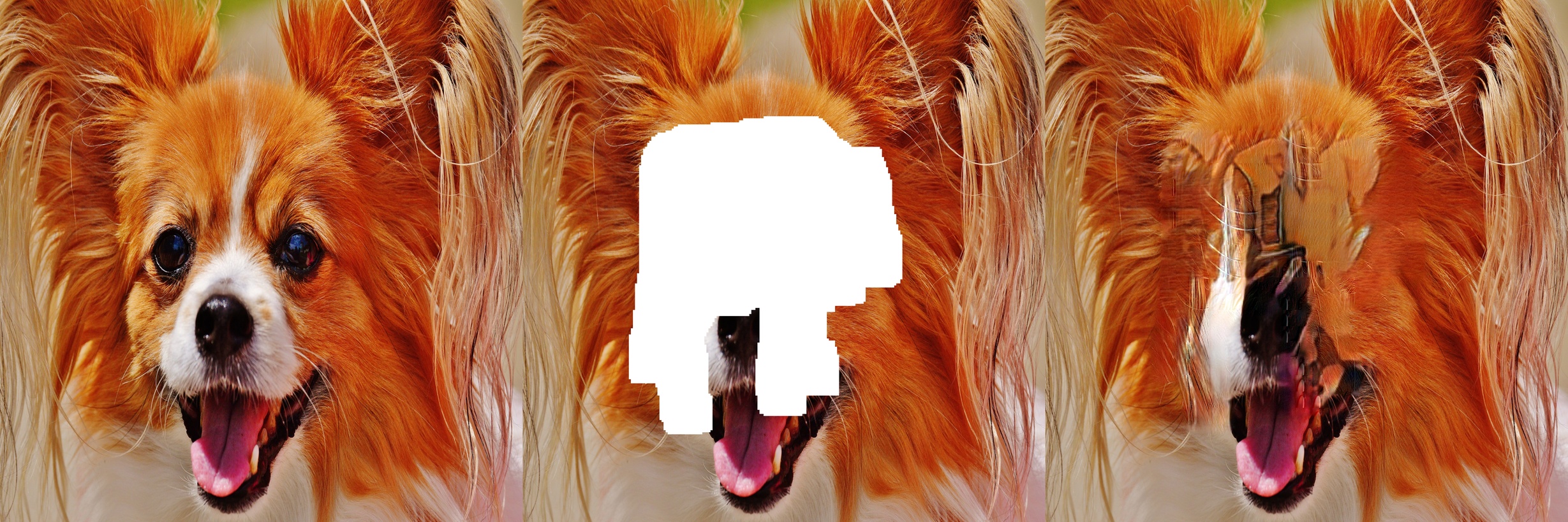}
\subfloat[Original]{\hspace{.17\linewidth}}
\subfloat[Input]{\hspace{.16\linewidth}}
\subfloat[Output]{\hspace{.17\linewidth}}
\subfloat[Original]{\hspace{.17\linewidth}}
\subfloat[Input]{\hspace{.16\linewidth}}
\subfloat[Output]{\hspace{.17\linewidth}}
\caption{Failure examples of our model.}
\label{fig:failure}
\end{center}
\end{figure*}

Some failure examples of our model are presented in Figure \ref{fig:failure}. Our model is prone to fail when the majority parts of a background object are missing (Referring to the bicycle and dog face in Figure. 4). 

\end{document}